\documentclass[runningheads]{llncs}
\usepackage{hyperref}
\usepackage{tabularx}
\usepackage[T1]{fontenc} 
\usepackage{graphicx}
\begin{document}
\title{Feature Selection Methods for Uplift Modeling and Heterogeneous Treatment Effect}
%
%
\author{Zhenyu Zhao\inst{1} \and
Yumin Zhang\inst{2} \and
Totte Harinen\inst{3} \and 
Mike Yung \inst{4}
}

\authorrunning{Z. Zhao et al.}
%
\institute{Tencent \email{zzy287@gmail.com} \and
Purdue University \email{zhan2013@purdue.edu} \\ \and
Toyota Research Institute \email{th.harinen@gmail.com} \and
Spotify \email{yungmsh@gmail.com}}  
\maketitle              
\begin{abstract}
Uplift modeling is a causal learning technique that estimates subgroup-level treatment effects. It is commonly used in industry and elsewhere for tasks such as targeting ads. In a typical setting, uplift models can take thousands of features as inputs, which is costly and results in problems such as overfitting and poor model interpretability. Consequently, there is a need to select a subset of the most important features for modeling. However, traditional methods for doing feature selection are not fit for the task because they are designed for standard machine learning models whose target is importantly different from uplift models. To address this, we introduce a set of feature selection methods explicitly designed for uplift modeling, drawing inspiration from statistics and information theory. We conduct empirical evaluations on the proposed methods on publicly available datasets, demonstrating the advantages of the proposed methods compared to traditional feature selection. We make the proposed methods publicly available as a part of the \textit{CausalML} open-source package.
\end{abstract}

\section{Introduction}
Uplift modeling \cite{chen2022imbalance,Grimmer2017-rl,guelman2015uplift,Gutierrez2016-co,Kunzel2017-ko,mouloud2020adapting,olaya2020survey,Rzepakowski2012-br,Soltys2015-be,teinemaa2021uplift,Wager2015-sd,Zaniewicz2013-rt,Zhao2017-kg,zhao2019uplift}, also known as heterogeneous treatment effect estimation or incremental modeling, is a technique designed to estimate the individual treatment effect (ITE) or the conditional average treatment effect (CATE) of an intervention. It is often used for user targeting, budget allocation, and personalization applications, including customer relationship management, promotion and incentives, advertisement, customer service,  recommendation systems, and product design. Such optimization is achieved by first estimating the ITE/CATE of the treatment for a user or group of users, and then delivering the optimal treatment strategy by considering the potential lift by and cost of the treatment. For example, it is often desired to target a product advertisement to users on whom the ads are likely to lift purchase conversion (a positive treatment effect) and skip users on whom the ads cannot make a difference on purchase conversion (a treatment effect close to $0$).

In practice, there is often a rich set of features that can be used to build an uplift model. However, using all of the available features in the model can lead to computational inefficiency, over-fitting, high maintenance workload, and model interpretation challenges. A feature selection method screens the large feature space and picks the important features for the model, and then the uplift model can be built based on the most important features. The feature selection step is beneficial for uplift modeling in multiple aspects: (1) faster computation speed for model training with fewer features; (2) more robust prediction by avoiding over-fitting; (3) lower maintenance cost for data pipelines; and (4) easier model interpretation and diagnostics. 
As Radcliffe and Surry \cite{radcliffe2011real} noted, feature selection is actually of greater importance to improve model quality and stability in uplift modeling than in conventional modeling because uplift models estimate the difference between two outcomes (considered as a second-order problem), which is often small relative to the direct outcomes (which traditional models estimate), increasing the risk of over-fitting markedly.

Feature selection methods for conventional machine learning problems have been well studied \cite{bolon2013review,chandrashekar2014survey,tang2014feature}. Nevertheless, as this paper will show, these methods are ineffective for solving feature selection problems for uplift modeling. The main reason is that such methods try to select predictive features of the outcome variable, which may not be the features related with treatment effect variability.

This paper assumes the data for uplift modeling is collected from a randomized experiment (a.k.a A/B testing \cite{kohavi2020trustworthy}), which is often the case in practice. 
It is essential to differentiate the feature selection methods for uplift modeling based on data from randomized experiments and the feature selection methods developed for observational causal inference. Feature selection algorithms for observational causal inference, such as the lasso-based approach proposed by \cite{shortreed2017outcome}, are designed to help models whose goal is to reduce confounding. These methods are out of scope for discussion in this paper as the framework and goals are different.

This work contributes to the feature selection of uplift modeling from both methodological and empirical perspectives, specifically:
\begin{itemize}
    \item Five filter methods and two types of embedded methods are proposed as feature selection methods for uplift modeling, each with its advantages, enriching the solution space for solving this problem.
    \item These feature selection methods are empirically evaluated and compared with ordinary feature selection methods as benchmarks on two data sets. One synthetic dataset is generated for this study and made available online \cite{zhao_zhenyu_2022_6342553}, and the other dataset is from Megafon Uplift Competition training data \cite{megafon2021}. 
    \item This study demonstrates logically and empirically that the ordinary feature selection methods for conventional machine learning models are sub-optimal in the uplift modeling context. 
    \item To make these feature selection methods easily accessible for broad applications, the proposed methods are implemented in the \textit{CausalML} \cite{chen2020causalml} open-source package. 
\end{itemize}

It is worth noting that the proposed feature selection methods are not only useful for uplift modeling purpose, but also useful for studying HTE in general. It can serve as an exploratory analysis of the feature space to narrow down the set of features causing HTE \cite{kl2015}. Such insights can identify pain points for specific user cohorts and iterate treatment intervention designs.  

To simplify the discussion, the rest of this paper assumes the outcome variable to be binary, covering most common use cases such as advertisement click-through, new user conversion, and existing user retention. All methods can be generalized to multi-class outcome scenarios. Some proposed methods (F filter, all embedded methods) can be generalized to regression problems for uplift modeling where the outcome variable is continuous.

The structure of the paper is as follows. 
Section \ref{background} reviews the critical concepts of uplift modeling and describes why feature selection for uplift modeling is a unique challenge. Section \ref{methodology} introduces five filter methods and three embedded methods for uplift modeling. 
Section \ref{examples} evaluates these methods with benchmark methods on two data sets. 
Finally, Section \ref{conclusion} summarizes the findings and makes recommendations for choosing and using the proper feature selection methods for uplift modeling applications.

\section{Background} \label{background}
\subsection{Uplift Models}
Uplift modeling can be viewed as a way to predict individual treatment effects using machine learning models. When there is heterogeneity in treatment effects, it means that ITE can be different from the ATE (average treatment effect), where the opportunity rises for serving users with different treatment based on their ITE (instead of applying control or treatment to all users).

It is helpful to frame the problem and introduce uplift modeling from a causal inference perspective. Following the commonly used Neyman-Rubin causal model \cite{Holland1986-dw,Neyman1923-kb,Rubin1974-xa,rubin2005causal}, the ITE for unit $i \in \{1,2,...,N\}$ can be expressed as:
\begin{equation}
    Y_i(1) - Y_i(0)
\end{equation}
where $Y_i(1)$ and $Y_i(0)$ denotes the outcome variable for individual $i$ under treatment condition and control condition respectively.

 A closely related concept is the conditional average treatment effect (CATE), which is the treatment effect for subgroups defined by combinations of covariate values. Let $X = (X_1, X_2,..., X_M)$ denote $M$ covariates (also called features) and $x_i = (x_{i1}, x_{i2},..., x_{iM})$ denote the observed realisation of $X$ for unit $i$. The conditional average treatment effect (CATE) is the expected treatment effect within groups of units that have a similar realisations of $X$:
\begin{eqnarray}
    \tau(x) := E[Y(1) - Y(0) \mid X = x]
\end{eqnarray}
Thus, CATE quantifies how treatment effects vary among units depending on the observed features. In practice, many uplift modeling approaches train on CATEs. \cite{Gutierrez2016-co}

There are two main categories of uplift models. 
The first category is known as meta-learners (\cite{Nie2017-uz,Kunzel2017-ko}). The models in this category are built based on conventional machine learning models as base learners in such a way that they can estimate CATEs. For example, the ``\textit{TwoModel}'' approach (\cite{Hansotia2001-vf}), also known as \textit{T-Learner}, is constructed by fitting a separate model for the control and treatment observations and then taking the difference between the predicted treatment outcome and the predicted control outcome to estimate the CATE. More complex meta-learners include \textit{X-Learner} and \textit{R-Learner} proposed by \cite{Kunzel2017-ko} and \cite{Nie2017-uz}, respectively. 

The other category of uplift models is known as uplift trees or causal trees. These models are based on modifying the loss functions within the classification or regression trees \cite{Athey2016-on,Wager2015-sd,guelman2015uplift,Guelman2012-bx,Rzepakowski2012-br,Athey2015-jd}. For example, \cite{Rzepakowski2012-br} propose modifying the splitting criterion of a classification tree algorithm such that the split is optimized for maximizing the heterogeneity of treatment effects in the resulting child notes. In this paper, feature selection methods are evaluated with models from both uplift modeling categories to test the generality of the proposed feature selection methods. 

\subsection{Related Work}
Several feature selection methods for uplift modeling purposes have been discussed in the literature. A filter method named net information value (NIV) \cite{kl2015} is built based on the net weight of evidence (NWOE), where NWOE is the difference between the weight of evidence (WOE) of treatment and WOE of control, and NIV is a weighted average of NWOE. 
Radcliffe \cite{radcliffe2011real} introduced the pessimistic qini estimate that uses one feature to train an uplift model at a time, and the resulting model accuracy measure (qini) is taken as the feature importance score. Certain regularizations are added to this method to improve stability. The pessimistic qini method can be more computationally costly as it involves training and potentially tuning an uplift model.

\subsection{Why Ordinary Feature Selection Methods Fail for Uplift Modeling}
There are various feature selection methods available for conventional classification and regression problems. The methods can be roughly divided into three categories: filter methods, wrapper methods, and embedded methods (\cite{bolon2013review,chandrashekar2014survey,tang2014feature}). However, these ordinary methods are unsuitable for selecting important uplift modeling features.

To help understand the reason, first notice the goals are different between conventional models and uplift models. Conventional models aim to predict the outcome variable ($Y$) while uplift models try to estimate the difference between two outcomes, one under the control condition and one under the treatment condition. An important feature for a conventional model is one that is predictive of the outcome variable, while an important feature for an uplift model is one that is predictive of the treatment effect.

As an example, consider a hypothetical randomized experiment in which social media influencers talk about a certain product in targeted advertisements. The advertiser is interested in measuring the ad's effect on customer conversion. If the social media influencers are not well known to older customers, we would expect age to predict treatment effect heterogeneity, assuming that the ad works:
\begin{eqnarray}
    E[Y(1) - Y(0) \mid Age = younger] > E[Y(1) - Y(0) \mid Age = older]
\end{eqnarray}
Even if age was the best predictor of treatment effect heterogeneity, it does not follow that it is the best predictor of whether customers convert or not. In extreme cases, age might be completely irrelevant for conversion in the absence of a treatment assignment, so that $E[Y \mid Age = younger] = E[Y \mid Age = older]$.

\section{Feature Selection Methods for Uplift Modeling} \label{methodology}
To identify the important features for uplift modeling, we propose both filter methods, which are easy and fast to use as a screening step for the data, and embedded methods, which are a by-product from training an uplift model. Each method can produce an importance score for each feature, and F filter and LR filter methods can additionally output p-values and statistical significance results for feature importances.

\subsection{Filter Methods}
As discussed above, a feature's importance in an uplift modeling task depends on how well it predicts the treatment effect. A filter method calculates the importance score for each feature based on the marginal relationship between the treatment effect and the feature. It is a fast pre-processing step because no uplift model training is involved, and it can be used for quickly screening the feature space and selecting important features for downstream modeling tasks.

\subsubsection{F Filter}
The F filter method is named after the F statistic for testing the significance of the interaction between the treatment indicator and a feature in linear regression. To capture possible nonlinear associations between features and the treatment effect, we extend this method by including higher-order terms of the features in the regression.
For a given feature $X_{j}$ ($j \in \{1,...,M\}$), the heterogeneous treatment effect of $X_{j}$ can be studied in the following regression:

\begin{eqnarray} \label{eq:1}
    Y = \alpha + \delta I + \sum_{r=1}^{R}\beta_r X_{j}^{r} + \sum_{r=1}^{R} \theta_r I X_{j}^{r} + \epsilon 
\end{eqnarray}
where $Y$ stands for the response variable, $I$ is the treatment indicator ($1$ for treatment, and $0$ for control), $R$ is a hyperparameter controlling the order to be studied ($X_{j}^r$ is $X_{j}$ to the power of $r$), $\epsilon$ represents the error term, and $\alpha$, $\delta$, $\beta$, $\theta$ are the coefficients, where the significance of $\theta$ indicates the strength of the heterogeneous treatment effect in the dimension of $X_{j}$.

The significance of $\theta$ can be studied by contrasting the model above with the following reduced model:
\begin{eqnarray} \label{eq:2}
    Y = \alpha' + \delta' I + \sum_{r=1}^{R}\beta'_r X_{j}^{r} + \epsilon' 
\end{eqnarray}

The feature importance score by F filter method is defined as the F-statistic for the coefficient of the interaction term: 
\begin{eqnarray}
    F = \frac{(RSS - RSS')/R}{RSS'/(N-R-2)}
\end{eqnarray}
with 
\begin{eqnarray*}
    RSS &=& \sum_{i=1}^{N}(y_i - \hat{\alpha} - \hat{\delta} I_i - \sum_{r=1}^{R}\hat{\beta}_r x_{ij}^{r} - \sum_{r=1}^{R} \hat{\theta}_r I x_{ij}^{r}) \\ 
    RSS' &=& \sum_{i=1}^{N}(y_i - \hat{\alpha'} - \hat{\delta'} I_i - \sum_{r=1}^{R}\hat{\beta}'_r x_{ij}^{r})
\end{eqnarray*}
where $N$ is the total number of observations, $RSS$ is the Residual Sum of Squares for fitted model \ref{eq:1} and $RSS'$ is the Residual Sum of Squares for fitted model \ref{eq:2}, that are calculated by plugging in the sample data and fitted coefficients. 

This F statistic follows an F distribution with $(R, N-R-2)$ degrees of freedom assuming that the true value of $\theta$ equals $0$. A byproduct of the F filter method is a p-value for the correlation between the feature and the treatment effect, which can be used to deem whether heterogeneity in a given feature dimension counts as statistically significant. Furthermore, statistical significance can be used as a cut-off rule for selecting important features. A straightforward extension of the F filter method is to calculate the importance and the significance of a set of multiple features at once: whether a set of features are associated with the heterogeneous treatment effect or not. The F filter method works for both the regression uplift modeling problem where $Y$ is continuous and the classification uplift modeling problem where $Y$ is categorical.

Setting the hyperparameter $R>1$ will enable the F filter to capture the nonlinear relationship between feature and HTE. As empirical results show in Section \ref{examples}, F filter with $R=2$ outperforms F filter with $R=1$ remarkably.

The F filter method can be directly applied to binary and continuous outcome, as well as continuous features. 

\subsubsection{LR Filter}
The LR (Likelihood Ratio) filter defines the feature importance score as the likelihood ratio test statistic for the interaction coefficient in a logistic regression model. Similar to the F filter, the LR filter for any feature $X_j$ is constructed by contrasting two logistic regression models:
\begin{eqnarray} \label{eq:lr1}
    \ \ logit(p(X_j,I;\alpha, \delta, \beta_1,...,\beta_R, \theta_1,...,\theta_R)) &=& \alpha + \delta I + \sum_{r=1}^{R}\beta_r X_j^r + \sum_{r=1}^{R}\theta_r I X_j^r \\
    \ \ logit(p(X_j,I;\alpha', \delta', \beta_1',...,\beta_R')) &=& \alpha' + \delta' I + \sum_{r=1}^R \beta_r' X_j^r
\end{eqnarray}
where $p(X_j,I;\alpha, \delta, \beta_1,...,\beta_R, \theta_1,...,\theta_R)$ and $p(X_j,I;\alpha', \delta', \beta_1',...,\beta_R')$ are probability representations of $Pr(Y=1|X_j,I)$ under two functions, and $R$ is a hyperparameter.

The significance of interaction coefficient $\theta$ can be tested through a likelihood ratio test. Let $\hat{\alpha}, \hat{\delta}, \hat{\beta_1},...,\hat{\beta_R}, \hat{\theta_1},...,\hat{\theta_R}, \hat{\alpha'}, \hat{\delta'}, \hat{\beta_1'},...,\hat{\beta_R'}$ denote the fitted coefficient estimates for the logistic regression models. The likelihood ratio statistic can be calculated by plugging in the sample data and fitted coefficients:
\begin{eqnarray} \label{eq:lr2}
    LR = &-& 2\sum_{i=1}^{N} [y_i log(p(x_{ij},I_i; \hat{\alpha}, \hat{\delta}, \hat{\beta_1},...,\hat{\beta_R}, \hat{\theta_1},...,\hat{\theta_R})) \\
    &+& (1-y_i) log(p(x_{ij},I_i; \hat{\alpha}, \hat{\delta}, \hat{\beta_1},...,\hat{\beta_R}, \hat{\theta_1},...,\hat{\theta_R})) \\ 
    &-& y_i log(p(x_{ij},I_i;\hat{\alpha'}, \hat{\delta'}, \hat{\beta_1'},...,\hat{\beta_R'})) \\
    &-& (1-y_i) log(p(x_{ij},I_i;\hat{\alpha'}, \hat{\delta'}, \hat{\beta_1'},...,\hat{\beta_R'})]
\end{eqnarray}

The LR statistic is taken as the feature importance score by the LR filter. Assuming that the true value of $\theta$ is 0, then the LR statistic follows a $\chi^2_{R}$ distribution. This LR filter method can also produce p-values for feature importances and be generalized to test the importance of a set of features. Similar to the F filter, setting $R>1$ for the LR filter extends its capability for capturing the nonlinear importance of features for HTE.

The LR filter method can be used for categorical outcome and continuous features. 

\subsubsection{Bin-based Divergence Filter (KL Filter, ED Filter, Chi Filter)}
The section introduces three variants of the bin-based divergence filter method, which are direct applications from the split criteria for uplift trees proposed by \cite{Rzepakowski2012-br}. Similar to the NIV method \cite{kl2015}, for a given feature, the bin-based method first divides the samples into $S$ (preferably equally sized) bins, where $S$ is a hyperparameter. The importance score is defined as the divergence measure of the treatment effect over these $S$ bins. 

Formally, let $p_{k}$ and $q_{k}$ denote the sample proportion of class $Y=1$ in the $k$th ($k = 1, \ldots, K$) bin for the treatment group and control group respectively. The KL (Kullback-Leibler divergence) filter, ED (squared Euclidean Distance), and Chi (chi-squared divergence) filter feature importance scores are defined as follows:
\begin{eqnarray} \label{divergence}
\Delta_{KL} &:=&  \sum_{k=1}^{K} \frac{N_{k}}{N} KL(p_k, q_k)= \sum_{k=1}^{K} \frac{N_{k}}{N} [p_{k} \log \frac{p_{k}}{q_{k}} + (1-p_{k}) \log \frac{1-p_{k}}{1-q_{k}} ] \\
\Delta_{ED} &:=& \sum_{k=1}^{K} \frac{N_{k}}{N} ED(p_k, q_k)= 2\sum_{k=1}^{K} \frac{N_{k}}{N} (p_{k} - q_{k})^{2} \\
\Delta_{Chi} &:=& \sum_{k=1}^{K} \frac{N_{k}}{N} \chi^2(p_k, q_k)=  \sum_{k=1}^{K} \frac{N_{k}}{N} [\frac{\left(p_{k} - q_{k}\right)^{2}}{q_{k}} + \frac{\left(p_{k} - q_{k}\right)^{2}}{1-q_{k}}]
\end{eqnarray}
where $N_k$ is the sample size in the $k$th bin.

Even though these bin-based filter methods share the same divergence measures as an uplift tree, they are simpler to implement and compute than training an uplift model. For data with $M$ features, the time complexity of the bin-based filter methods is linear with sample size, $O(M \cdot N)$, while the time complexity of building a single complete uplift tree is $O(M \cdot N \log(N))$,

The bin-based divergence filter method can be directly applied to categorical outcome and continuous features. For discrete features, this method can be applied without binning. 

\subsection{Embedded Methods}
The final category of methods that we propose, the embedded methods, obtain feature importances as a byproduct of training an uplift model. The way in which the importance scores are calculated differs depending on whether the target model is a meta-learner or an uplift tree. The embedded methods can be used for continuous and categorical outcomes, as well as continuous and discrete features. 

\subsubsection{Embedded Methods by Meta-learners}
For meta-learners, feature importance can be obtained from the base-learners, which are the composite models making up a meta-learner.

The simplest meta-learner, the \textit{OneModel} or \textit{S-Learner}, is based on just one base-learner and predicts treatment effects as the difference between $E[Y \mid X=x, I=1]$ and $E[Y \mid X=x, I=0]$ where $I$ is the treatment indicator. Because the importance scores from the single base-learner in \textit{S-Learner} tend to be similar to an outcome based model (the only difference between the two is that the base-learner in \textit{S-Learner} has a treatment indicator as an additional feature), we will not differentiate these two embedded methods and let \textit{Outcome} embedded represent these two methods in the following discussion.

The \textit{TwoModel} embedded feature selection method is derived from the \textit{TwoModel} uplift model, which feature importance score is defined as the sum of its embedded importance scores produced by the two base-learners.  As the embedded methods derived from Meta-learners are based on ordinary feature selection methods in base learners (conventional model), the feature selection performance for uplift modeling is expected to be poor as it does not consider HTE during the process.

\subsubsection{Embedded Methods by Uplift Trees}
For uplift trees, the importance score for a feature can be defined as the cumulative contribution to the loss function during the tree node splits in the trees. This is similar to the well-known embedded feature importance for standard classification trees, except the score is obtained from an uplift tree with a special splitting criterion. At each split, we calculate the gain in the distribution divergence:
\begin{equation}
    \Delta = \sum_{k \in \mbox{\{left, right\}}} \frac{n_k}{n} D\left(p_{k}, q_{k}\right) - D(p, q),
\end{equation}
where $n$ is the sample size in the parent node, $n_k$ is the sample size in the child node, $D()$ is divergence measure defined as $KL(), ED(), \chi^2()$ in Eq.(~\ref{divergence}), $p$ and $q$ denote the proportion of $Y=1$ for the treatment group and control group separately in the parent node, and $p_k$, $q_k$ are corresponding proportions in the child notes. The feature importance score is calculated by summing over all the $\Delta$ from the tree node splits where the feature is used.

The time complexity for embedded methods depend on the learners used. For random forest algorithms, it is at order of $O(t_{tree} \cdot m_{s} \cdot N \cdot \log(N))$ where $t_{tree}$ is the number of trees and $m_{s}$ is the maximum features considered in each split.

\section{Empirical Evaluation} \label{examples}
This section aims to answer the following questions: (1) Which feature selection method works better than others empirically? (2) Is feature selection performance consistent and generalizable with different uplift models? (3) How does the feature selection step affect the accuracy of uplift modeling? The approach for evaluating the performance of a feature selection method is to feed the top features selected by this method to an uplift model and assess the accuracy of the uplift model output. 

\subsection{Experiment 1: Evaluation with Synthetic Data}
\begin{figure*}[h]
\centering
\includegraphics[width=1. \textwidth]{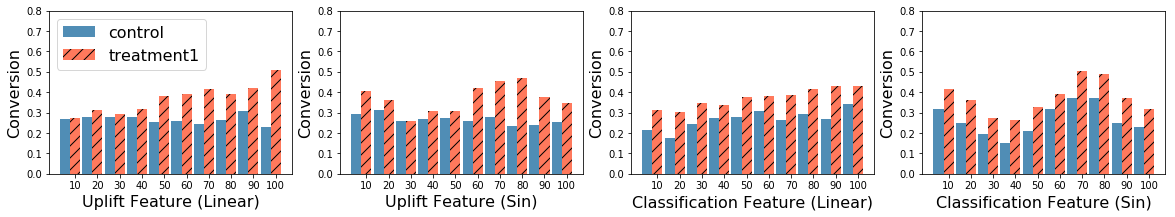}
\caption{Feature Association Pattern with Outcome by Experiment Group in Experiment 1. The first two plots demonstrate a heterogeneous treatment effect associated with uplift features in a linear and sine pattern respectively. The last two plots illustrate classification features are correlated with outcome, but not treatment effect.}
\label{fig:feature_pattern}
\end{figure*}

For evaluating the feature selection methods, a specific synthetic data generation process is designed, such that the data contains three types of features: (1) uplift features influencing the treatment effect on the conversion probability; (2) classification features affecting the conversion probability but independent of the treatment effect; and (3) irrelevant features that are independent of both conversion probability and the treatment effect. 

The binary response variable is generated based on the probability $Pr(Y = 1 | X, I)$, where $X$ denotes the feature vector and $I$ is the treatment indicator. Assuming there are $m_1$ classification features and $m_2$ uplift features, the probability $Pr(Y = 1 | X, I)$ itself is modeled as:
\begin{eqnarray*}
logit(Pr(Y=1 | X, I)) = a_1 + \sum_{j=1}^{m_1} f_j(X_{j}) \beta_j  + I \cdot (a_2 + \sum_{j=m_1+1}^{m_1+m_2} f_j(X_{j}) \beta_j) + \epsilon
\end{eqnarray*}
where $a_1, a_2, \{\beta_j\}_{1}^{m1+m2}$ are coefficient parameters, $\epsilon$ is a random noise added,  $\{f_j(\cdot)\}_{1}^{m1+m2}$ is transformation function for the features, such that features have six types of association with the outcome: linear, quadratic, cubic, ReLU (Rectified Linear Unit \cite{dahl2013improving}), trigonometric function sine, and cosine. Example feature patterns are plotted in Figure ~\ref{fig:feature_pattern}. 

In this study, there are $36$ features in total, including $m_1 = 10$ classification features, $m_2 = 6$ uplift features, and $m_3 = 20$ irrelevant features. The values for the coefficients are set such that the average control conversion probability is around $0.2$ and the average treatment effect is around $0.1$.
The generated synthetic data is published online \cite{zhao_zhenyu_2022_6342553}.

Eleven feature selection methods are evaluated, including six uplift filter methods (F filter, LR filter, KL filter, Chi filter, ED filter, NIV filter), two uplift embedded methods (\textit{TwoModel} embedded and KL embedded), two ordinary filter methods (ordinary Mutual Information and ordinary F Score \cite{scikit-learn}), and one ordinary embedded method \textit{Outcome embedded} as a benchmark \cite{scikit-learn}. For F filter and LR filter, variants (F, F2, F3, LR, LR2, LR3) are created based on $R=1,2,3$. 
For the bin-based filter methods, the number of bins is set at $10$ as default. Four uplift models are used to evaluate the performance of the feature selection methods: \textit{TwoModel}, \textit{X-Learner}, \textit{R-Learner}, and uplift random forest (with KL criterion), given the results are similar in the order of feature selection method performance, only \textit{TwoModel} and uplift random forest results are presented in the paper. For all the meta-learners, a random forest classifier ($n\_estimators = 10, max\_depth=10, min\_child\_samples=100$) \cite{scikit-learn} is used as the base learner. 

Each simulation trial consists of four steps. First, the data generator simulates the data with a new random seed and randomly splits the data into training and testing (with $50\%:50\%$ ratio). Second, each feature selection method is applied to the training data and ranks the features by importance score. 
Third, for each feature selection method, the top $m^*$ (for $m^* \in \{2,4,6,8,10\}$) features are used to build uplift models using training data. 
Fourth, the testing data is used to evaluate the accuracy of the uplift models based on the top features selected by each feature selection method. 
Each trial generates $20,000$ samples ($10,000$ for control and $10,000$ for treatment) and the simulation study consists of $t=50$ trials. 

\begin{table*}[h]
\begin{center}
\caption{RMSE (and Standard Deviation of RMSE) for estimating ITE using Uplift Random Forest as evaluating model based on Synthetic Data (RMSE lower the better).}
\label{table:syn_data_rmse_kl}
\begin{tabular}{|c|c|c|c|c|c|}
\hline
Top Features & 2& 4 & 6 & 8 & 10 \\
\hline 
F filter & 0.184(0.004) & 0.173(0.006) & 0.172(0.006) & 0.173(0.006) & 0.172(0.007) \\
F2 filter & 0.183(0.003) & 0.158(0.003) & 0.160(0.004) & 0.162(0.005) & 0.163(0.005) \\
F3 filter & 0.184(0.003) & 0.159(0.003) & 0.159(0.005) & 0.160(0.005) & 0.163(0.005) \\
LR filter & 0.184(0.004) & 0.174(0.006) & 0.172(0.006) & 0.173(0.005) & 0.172(0.007) \\
LR2 filter & 0.183(0.003) & 0.158(0.003) & 0.160(0.004) & 0.162(0.004) & 0.164(0.004) \\
LR3 filter & 0.184(0.003) & 0.159(0.003) & 0.160(0.004) & 0.161(0.004) & 0.162(0.004) \\
KL filter & 0.185(0.003) & 0.162(0.004) & 0.148(0.004) & 0.151(0.003) & 0.155(0.003) \\
ED filter & 0.185(0.003) & 0.161(0.004) & 0.146(0.004) & 0.150(0.004) & 0.155(0.004) \\
Chi filter & 0.185(0.003) & 0.162(0.004) & 0.149(0.005) & 0.152(0.004) & 0.156(0.003) \\
NIV filter & 0.185(0.003) & 0.162(0.004) & 0.147(0.004) & 0.151(0.003) & 0.155(0.004) \\
KL embedded & 0.186(0.004) & 0.167(0.011) & 0.159(0.009) & 0.159(0.008) & 0.161(0.007) \\
TwoModel embedded & 0.208(0.006) & 0.195(0.010) & 0.184(0.009) & 0.174(0.009) & 0.167(0.007) \\
Outcome embedded & 0.210(0.006) & 0.204(0.009) & 0.193(0.011) & 0.181(0.011) & 0.173(0.011) \\
Ordinary FScore & 0.207(0.007) & 0.193(0.011) & 0.180(0.009) & 0.174(0.004) & 0.173(0.006) \\
Ordinary MutualInfo & 0.208(0.007) & 0.204(0.010) & 0.197(0.010) & 0.192(0.012) & 0.189(0.011) \\
\hline
\end{tabular}
\end{center}
\end{table*}

\begin{table*}
\begin{center}
\caption{RMSE (and Standard Deviation of RMSE) for estimating ITE using \textit{TwoModel} as evaluating model based on Synthetic Data (RMSE lower the better).}
\label{table:syn_data_rmse_twomodel}
\begin{tabular}{|c|c|c|c|c|c|}
\hline
Top Features & 2& 4 & 6 & 8 & 10 \\
\hline 
F filter & 0.183(0.005) & 0.174(0.005) & 0.174(0.006) & 0.174(0.006) & 0.174(0.007) \\
F2 filter & 0.184(0.004) & 0.163(0.003) & 0.165(0.005) & 0.166(0.006) & 0.168(0.005) \\
F3 filter & 0.185(0.004) & 0.163(0.003) & 0.163(0.005) & 0.165(0.006) & 0.167(0.005) \\
LR filter & 0.184(0.005) & 0.175(0.006) & 0.174(0.006) & 0.176(0.006) & 0.175(0.006) \\
LR2 filter & 0.184(0.004) & 0.163(0.003) & 0.166(0.005) & 0.168(0.006) & 0.170(0.005) \\
LR3 filter & 0.186(0.004) & 0.163(0.003) & 0.165(0.005) & 0.167(0.005) & 0.168(0.005) \\
KL filter & 0.184(0.004) & 0.163(0.005) & 0.153(0.006) & 0.162(0.006) & 0.165(0.005) \\
ED filter & 0.184(0.003) & 0.162(0.004) & 0.151(0.004) & 0.157(0.006) & 0.160(0.005) \\
Chi filter & 0.184(0.004) & 0.163(0.005) & 0.154(0.007) & 0.162(0.006) & 0.165(0.005) \\
NIV filter & 0.184(0.004) & 0.163(0.005) & 0.152(0.006) & 0.158(0.005) & 0.162(0.005) \\
KL embedded & 0.187(0.005) & 0.171(0.011) & 0.165(0.009) & 0.165(0.008) & 0.167(0.007) \\
TwoModel embedded & 0.210(0.007) & 0.198(0.010) & 0.191(0.008) & 0.184(0.007) & 0.181(0.006) \\
Outcome embedded & 0.212(0.006) & 0.207(0.009) & 0.199(0.009) & 0.190(0.009) & 0.185(0.008) \\
Ordinary FScore & 0.208(0.008) & 0.195(0.011) & 0.185(0.008) & 0.181(0.003) & 0.181(0.006) \\
Ordinary MutualInfo & 0.211(0.006) & 0.207(0.009) & 0.202(0.010) & 0.198(0.010) & 0.196(0.010) \\
\hline
\end{tabular}
\end{center}
\end{table*}

\begin{table*}
\begin{center}
\caption{Proportion of Uplift Features Selected in Top $6$ Positions by Method (Feature Recall). The table is ranked by the 'All Uplift' column, that indicates proportion of all uplift features ($6$ in total) being captured in the top $6$ features ranked by each method. A breakdown of feature recall score by different uplift feature pattern is presented.}
\label{table:syn_data_feature_recall}
\begin{tabular}{|c|c|c|c|c|c|c|c|}
\hline 
Method  & All Uplift & Linear  & Quadratic  & Cubic & ReLU & Sin & Cos \\
\hline
F filter & 55\% & 100\% & 12\% & 100\% & 100\% & 6\% & 12\% \\
F2 filter & 69.3\% & 100\% & 100\% & 100\% & 100\% & 4\% & 12\%  \\
F3 filter & 70.3\% & 100\% & 100\% & 100\% & 100\% & 10\% & 12\%  \\
LR filter & 55\% & 100\% & 10\% & 100\% & 100\% & 6\% & 14\%  \\
LR2 filter & 69.3\% & 100\% & 100\% & 100\% & 100\% & 6\% & 10\% \\
LR3 filter & 69.3\% & 100\% & 100\% & 100\% & 100\% & 6\% & 10\%  \\
KL filter & 97.7\% & 100\% & 100\% & 90\% & 100\% & 98\% & 98\%  \\
ED filter & 99.7\% & 100\% & 100\% & 98\% & 100\% & 100\% & 100\% \\
Chi filter & 95.3\% & 100\% & 100\% & 80\% & 98\% & 98\% & 96\% \\
NIV filter & 98.3\% & 100\% & 100\% & 90\% & 100\% & 100\% & 100\%  \\
KL embedded & 75.7\% & 94\% & 88\% & 82\% & 86\% & 42\% & 62\%  \\
TwoModel embedded & 32.7\% & 66\% & 8\% & 22\% & 92\% & 6\% & 2\%  \\
Outcome embedded & 23\% & 66\% & 20\% & 8\% & 42\% & 0\% & 2\%  \\
Ordinary FScore & 38.7\% & 96\% & 0\% & 46\% & 88\% & 2\% & 0\%  \\
Ordinary MutualInfo & 20\% & 24\% & 22\% & 14\% & 20\% & 12\% & 28\% \\
\hline
\end{tabular}
\end{center}
\end{table*}

A preferred feature selection method is expected to provide important features that lead to accurate ITE estimation by the downstream uplift model based on the features selected. 
Table \ref{table:syn_data_rmse_kl} and Table \ref{table:syn_data_rmse_twomodel} summarize the RMSE (Root Mean Square Error) of ITE estimation and its standard deviation over synthetic data trials using feature selection methods by uplift random forest (KL criterion) and \textit{TwoModel} respectively. 
As a benchmark, if all $36$ features are used in an uplift model, then the RMSE by uplift random forest is $0.181$ (with standard deviation $0.004$), and the RMSE by \textit{TwoModel} is $0.186$ (with standard deviation $0.003$).

The results show that 
1) the ordinary feature selection methods (ordinary F Score and ordinary Mutual Information) have poor performance and are not suitable for selecting top features for uplift modeling; 
2) the bin-based divergence filter methods (KL filter, ED filter, Chi filter, NIV filter) have consistent top performance in all scenarios, followed by KL embedded, and F2, F3, LR2, LR3 filters; 
3) adding higher-order terms in F filter and LR filter ($R>1$) improved the performance;
4) the outcome embedded method and the \textit{TwoModel} embedded method have essentially similar logic as ordinary feature selection methods, thus failing in the tasks. 
The relative performance of feature selection methods is consistent for different uplift models.

To better understand what features are selected by each method, we report the proportion of uplift features selected in top $6$ positions in Table ~\ref{table:syn_data_feature_recall}. The proportion is averaged across the $100$ trials. Note that there are $6$ uplift features out of $36$ in each trial, one in each pattern category. 
The detailed breakdown by uplift feature type explains why some methods are not performing well. F filter and LR filter fail to capture quadratic features, Sin features, and Cos features. Setting $R>1$ for these two methods improved the capability for capturing quadratic features. The ordinary filter methods and \textit{Outcome} embedded and \textit{TwoModel} embedded methods fail to capture uplift features in general.

\subsection{Experiment 2: Evaluation with MegaFon Uplift Competition Data}

\begin{figure*}[h]
\centering
\includegraphics[width=1\textwidth]{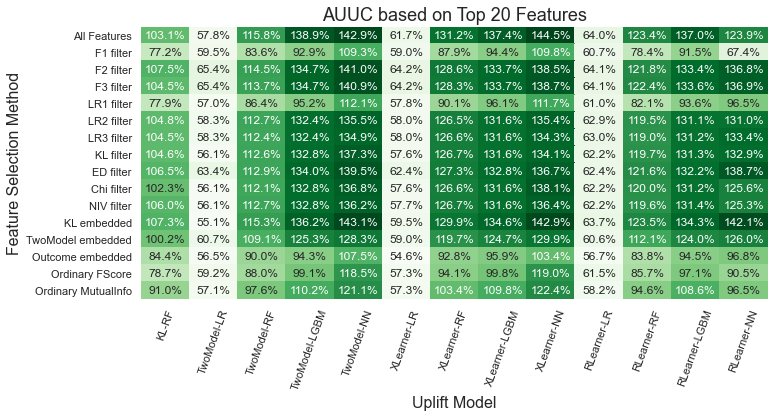}
\caption{AUUC of Uplift Models Using the Top $20$ Selected Features by Different Feature Selection Methods in the MegaFon Data Experiment with $300,000$ training data and $300,000$ testing data (AUUC larger the better). The Y-axis shows the feature selection method and the X-axis shows the uplift model (in a \{uplift model - base learner\} format) used for producing the AUUC score. The first row represents the AUUC using all $50$ features as reference.}
\label{fig:real_data_auuc}
\end{figure*}

\begin{figure*}
\centering
\includegraphics[width=1\textwidth]{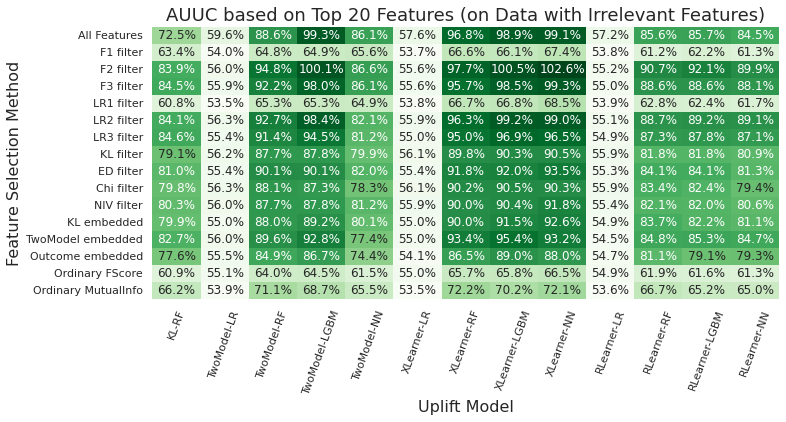}
\caption{AUUC of Uplift Models Using the Top $20$ Selected Features by Different Feature Selection Methods in the Noise-added MegaFon Data Experiment with $10,000$ training data and $10,000$ testing data (repeated for $30$ times). $50$ irrelevant features are added. The first row represents the AUUC using all $100$ features as reference.}
\label{fig:real_data_auuc2}
\end{figure*}

This example uses a data set that is publicly available for an online competition \cite{megafon2021}, that contains $600,000$ observations and $50$ features. 

To test the generality of the feature selection methods for different uplift models, the uplift models considered include: (1) \textit{TwoModel-LR}, \textit{X-Learner-LR}, \textit{R-Learner-LR} using \{ Logistic Regression Classifier $\&$ Linear Regression Regressor \} as base learners; (2) \textit{TwoModel-LGBM}, \textit{X-Learner-LGBM}, \textit{R-Learner-LGBM} using \{ Gradient Boosting Classifier $\&$ Gradient Boosting Regressor \} from \textit{LightGBM} implementation \cite{ke2017lightgbm} as base learners, with hyperparameter values ($n\_estimators = 100, max\_depth=10, min\_child\_samples=100$); (3) \textit{TwoModel-RF}, \textit{X-Learner-RF}, \textit{R-Learner-RF} using \{ Random Forest Classifier $\&$ Random Forest Regressor \} as base learners, with hyper-parameter values ($n\_estimators = 100, max\_depth=10, min\_samples\_leaf=100$); (4) \textit{TwoModel-NN}, \textit{X-Learner-NN}, \textit{R-Learner-NN} using \{ Neural Network (Multi-layer Perceptron) Classifier $\&$ Neural Network Regressor \}  from scikit-learn \cite{pedregosa2011scikit} implementation as base learners: each contains two layers with $100$ neurons on the first layer and $10$ neurons on the second layer; (5) \textit{KL-RF} as the uplift random forest using KL divergence criterion with hyper-parameter values ($n\_estimators = 20, max\_depth=10, min\_samples\_leaf=100$). 

The first evaluation of this data is conducted by equally splitting the data into training data and testing data, each with $300,000$ observations, where training data is used for feature selection and uplift modeling training, and testing data is used for evaluating the accuracy of uplift modeling. 
The results are summarized in Figure \ref{fig:real_data_auuc}, reporting the AUUC (area under the uplift curve) scores \cite{Soltys2015-be,Zhao2017-kg,Rzepakowski2012-br,Gutierrez2016-co} from the uplift models using the top $20$ features selected by each feature selection method. 
The relative performance of different feature selection methods can be compared within each column given the same uplift model.
The results are roughly aligned with the synthetic data results, except the performance of F2, F3, LR2, LR3 filters become comparable with other bin-based divergence methods. A possible reason is that this dataset does not have extreme nonlinear feature patterns like the sine or cosine forms in the previous synthetic data. Despite the differences in uplift models, the relative order of feature selection method performance is quite consistent across different uplift models. Models using all $50$ features still perform better than models using $20$ features. One reason is that the sample size is large compared with the feature space, in which case over-fitting is unlikely to occur. 

To test the feature selection methods in a more noisy setting, a second study is done by enriching the original dataset by adding $50$ irrelevant features (generated from a standard uniform distribution). The original data is divided into $30$ sub data set; each sub data set is further split into $10,000$ training sample and $10,000$ testing sample. All other model settings keep the same. The average AUUC over the $30$ trials are summarized in Figure \ref{fig:real_data_auuc2}. In this case, the sample to feature ratio becomes smaller, and the model is more likely to overfit. The results show the uplift models with $20$ important features selected can outperform uplift models with all $100$ features. F filter and LR filter with $R>1$ have the best performance in this experiment, indicating their robustness in noisy data.

\section{Discussion}
Our experiments demonstrate that the proposed methods can select important features based on their association with heterogeneous treatment effects and improve the ability of uplift models to estimate individual treatment effects. As shown above, ordinary feature selection methods perform poorly for uplift modeling tasks. The F filter and LR filter can enhance performance by adding nonlinear terms, and such methods can perform well in noisy data settings. The bin-based divergence filter methods also show good performance, although their performance is expected to suffer if the feature space makes it hard to create equal-sized bins. The embedded method with uplift random forest also perform competitively because the split criterion for such trees directly models HTE in child nodes. This approach, however, is more time-consuming compared to the other variable selection methods because it involves uplift model training. Additionally, its performance depends on the uplift model's hyperparameters.

\section{Conclusion} \label{conclusion}
We started by hypothesizing that traditional variable selection methods are not suitable for uplift models. We developed a number of alternative approaches that took into account the unique character of uplift models, namely the fact that the target in such models is the treatment effect. We then tested our proposed methods in a number of experiments, establishing that they conclusively outperform traditional variable selection approaches. We further highlighted different strengths and weaknesses of the proposed methods. 

Given the increasing use of uplift model in industry applications and elsewhere, we believe the proposed methods to prove practically relevant for several purposes, including accuracy, efficiency and interpretability. For this reason, we have made the proposed methods publicly available in the \textit{Causal ML} open source library. Promising areas for future development include covering more of the ever-increasing group of uplift models, applying the methods to observational in addition to experimental data, and integrating the methods with emerging model interpretability approaches.

The future work may include a detailed empirical study of feature selection methods for continuous outcome, as well as discussing feature selection methods for selecting the best set of features (collaboratively instead of individually). 

\bibliographystyle{splncs04}
\bibliography{uplift_references}

\begin{thebibliography}{10}
\providecommand{\url}[1]{\texttt{#1}}
\providecommand{\urlprefix}{URL }
\providecommand{\doi}[1]{https://doi.org/#1}

\bibitem{megafon2021}
{Megafon Uplift Competition} (2021),
  \url{https://ods.ai/tracks/df21-megafon/competitions/megafon-df21-comp/data}

\bibitem{Athey2015-jd}
Athey, S., Imbens, G.: Recursive partitioning for heterogeneous causal effects
  (Apr 2015)

\bibitem{Athey2016-on}
Athey, S., Tibshirani, J., Wager, S.: Generalized random forests  (Oct 2016)

\bibitem{bolon2013review}
Bol{\'o}n-Canedo, V., S{\'a}nchez-Maro{\~n}o, N., Alonso-Betanzos, A.: A review
  of feature selection methods on synthetic data. Knowledge and information
  systems  \textbf{34}(3),  483--519 (2013)

\bibitem{chandrashekar2014survey}
Chandrashekar, G., Sahin, F.: A survey on feature selection methods. Computers
  \& Electrical Engineering  \textbf{40}(1),  16--28 (2014)

\bibitem{chen2020causalml}
Chen, H., Harinen, T., Lee, J.Y., Yung, M., Zhao, Z.: Causalml: Python package
  for causal machine learning. arXiv preprint arXiv:2002.11631  (2020)

\bibitem{chen2022imbalance}
Chen, X., Liu, Z., Yu, L., Yao, L., Zhang, W., Dong, Y., Gu, L., Zeng, X., Tan,
  Y., Gu, J.: Imbalance-aware uplift modeling for observational data  (2022)

\bibitem{dahl2013improving}
Dahl, G.E., Sainath, T.N., Hinton, G.E.: Improving deep neural networks for
  lvcsr using rectified linear units and dropout. In: 2013 IEEE international
  conference on acoustics, speech and signal processing. pp. 8609--8613. IEEE
  (2013)

\bibitem{Grimmer2017-rl}
Grimmer, J., Messing, S., Westwood, S.J.: Estimating heterogeneous treatment
  effects and the effects of heterogeneous treatments with ensemble methods.
  Polit. Anal.  \textbf{25}(4),  413--434 (Oct 2017)

\bibitem{Guelman2012-bx}
Guelman, L., Guill{\'e}n, M., P{\'e}rez-Mar{\'\i}n, A.M.: Random forests for
  uplift modeling: An insurance customer retention case. In: Modeling and
  Simulation in Engineering, Economics and Management. pp. 123--133. Springer
  Berlin Heidelberg (2012)

\bibitem{guelman2015uplift}
Guelman, L., Guill{\'e}n, M., P{\'e}rez-Mar{\'i}n, A.M.: Uplift random forests.
  Cybernetics and Systems  \textbf{46}(3-4),  230--248 (2015)

\bibitem{Gutierrez2016-co}
Gutierrez, P., Gerardy, J.Y.: Causal inference and uplift modeling a review of
  the literature. JMLR: Workshop and Conference Proceedings 67  (2016)

\bibitem{Hansotia2001-vf}
Hansotia, B., Rukstales, B.: Incremental value modeling. Research Council
  Journal  (2001)

\bibitem{Holland1986-dw}
Holland, P.W.: Statistics and causal inference. J. Am. Stat. Assoc.
  \textbf{81}(396),  945--960 (1986)

\bibitem{ke2017lightgbm}
Ke, G., Meng, Q., Finley, T., Wang, T., Chen, W., Ma, W., Ye, Q., Liu, T.Y.:
  Lightgbm: A highly efficient gradient boosting decision tree. In: Advances in
  neural information processing systems. pp. 3146--3154 (2017)

\bibitem{kohavi2020trustworthy}
Kohavi, R., Tang, D., Xu, Y.: Trustworthy online controlled experiments: A
  practical guide to a/b testing. Cambridge University Press (2020)

\bibitem{Kunzel2017-ko}
K{\"u}nzel, S.R., Sekhon, J.S., Bickel, P.J., Yu, B.: Meta-learners for
  estimating heterogeneous treatment effects using machine learning  (Jun 2017)

\bibitem{kl2015}
Larsen, K.: Data exploration with weight of evidence and information value in r
   (2015)

\bibitem{mouloud2020adapting}
Mouloud, B., Olivier, G., Ghaith, K.: Adapting neural networks for uplift
  models. arXiv preprint arXiv:2011.00041  (2020)

\bibitem{Neyman1923-kb}
Neyman, J.: Sur les applications de la th{\'e}orie des probabilit{\'e}s aux
  experiences agricoles: Essai des principes. Roczniki Nauk Rolniczych
  \textbf{10},  1--51 (1923)

\bibitem{Nie2017-uz}
Nie, X., Wager, S.: {Quasi-Oracle} estimation of heterogeneous treatment
  effects  (Dec 2017)

\bibitem{olaya2020survey}
Olaya, D., Coussement, K., Verbeke, W.: A survey and benchmarking study of
  multitreatment uplift modeling. Data Mining and Knowledge Discovery
  \textbf{34}(2),  273--308 (2020)

\bibitem{scikit-learn}
Pedregosa, F., Varoquaux, G., Gramfort, A., Michel, V., Thirion, B., Grisel,
  O., Blondel, M., Prettenhofer, P., Weiss, R., Dubourg, V., Vanderplas, J.,
  Passos, A., Cournapeau, D., Brucher, M., Perrot, M., Duchesnay, E.:
  Scikit-learn: Machine learning in {P}ython. Journal of Machine Learning
  Research  \textbf{12},  2825--2830 (2011)

\bibitem{pedregosa2011scikit}
Pedregosa, F., Varoquaux, G., Gramfort, A., Michel, V., Thirion, B., Grisel,
  O., Blondel, M., Prettenhofer, P., Weiss, R., Dubourg, V., et~al.:
  Scikit-learn: Machine learning in python. Journal of machine learning
  research  \textbf{12}(Oct),  2825--2830 (2011)

\bibitem{radcliffe2011real}
Radcliffe, N.J., Surry, P.D.: Real-world uplift modelling with
  significance-based uplift trees. White Paper TR-2011-1, Stochastic Solutions
  pp. 1--33 (2011)

\bibitem{Rubin1974-xa}
Rubin, D.B.: Estimating causal effects of treatments in randomized and
  nonrandomized studies. J. Educ. Psychol.  \textbf{66}(5),  688--701 (1974)

\bibitem{rubin2005causal}
Rubin, D.B.: Causal inference using potential outcomes: Design, modeling,
  decisions. Journal of the American Statistical Association
  \textbf{100}(469),  322--331 (2005)

\bibitem{Rzepakowski2012-br}
Rzepakowski, P., Jaroszewicz, S.: Decision trees for uplift modeling with
  single and multiple treatments. Knowl. Inf. Syst.  \textbf{32}(2),  303--327
  (Aug 2012)

\bibitem{shortreed2017outcome}
Shortreed, S.M., Ertefaie, A.: Outcome-adaptive lasso: variable selection for
  causal inference. Biometrics  \textbf{73}(4),  1111--1122 (2017)

\bibitem{Soltys2015-be}
So{\l}tys, M., Jaroszewicz, S., Rzepakowski, P.: Ensemble methods for uplift
  modeling. Data Min. Knowl. Discov.  \textbf{29}(6),  1531--1559 (Nov 2015)

\bibitem{tang2014feature}
Tang, J., Alelyani, S., Liu, H.: Feature selection for classification: A
  review. Data classification: algorithms and applications p.~37 (2014)

\bibitem{teinemaa2021uplift}
Teinemaa, I., Albert, J., Goldenberg, D.: Uplift modeling: From causal
  inference to personalization (2021)

\bibitem{Wager2015-sd}
Wager, S., Athey, S.: Estimation and inference of heterogeneous treatment
  effects using random forests  (Oct 2015)

\bibitem{Zaniewicz2013-rt}
Zaniewicz, L., Jaroszewicz, S.: Support vector machines for uplift modeling.
  In: 2013 {IEEE} 13th International Conference on Data Mining Workshops. pp.
  131--138 (Dec 2013)

\bibitem{Zhao2017-kg}
Zhao, Y., Fang, X., Simchi-Levi, D.: Uplift modeling with multiple treatments
  and general response types  (May 2017)

\bibitem{zhao_zhenyu_2022_6342553}
Zhao, Z.: {Synthetic Data for Uplift Modeling and Heterogenous Treatment Effect
  with Known Counterfactuals and ITE} (Mar 2022),
  \url{https://doi.org/10.5281/zenodo.6342553}

\bibitem{zhao2019uplift}
Zhao, Z., Harinen, T.: Uplift modeling for multiple treatments with cost
  optimization. arXiv preprint arXiv:1908.05372  (2019)

\end{thebibliography}
\end{document}